\documentclass[10pt,twocolumn,letterpaper]{article}

\usepackage[pagenumbers]{cvpr} 

\usepackage{graphicx}
\usepackage{amsmath}
\usepackage{amssymb}
\usepackage{booktabs}
\usepackage{graphicx}
\usepackage{algorithm2e}
\usepackage{amsmath}
\usepackage{multirow}
\usepackage{multicol}
\usepackage{booktabs}
\usepackage{colortbl}
\usepackage{makecell}
\definecolor{mygray}{gray}{.9}
\definecolor{myblue}{rgb}{0.9, 0.9, 1}
\usepackage{svg}
\usepackage{booktabs}
\usepackage{marvosym}
\usepackage[numbers]{natbib}

\definecolor{cvprblue}{rgb}{0.21,0.49,0.74}
\usepackage[pagebackref,breaklinks,colorlinks,allcolors=cvprblue]{hyperref}

\usepackage[capitalize]{cleveref}
\crefname{section}{Sec.}{Secs.}
\Crefname{section}{Section}{Sections}
\Crefname{table}{Table}{Tables}
\crefname{table}{Tab.}{Tabs.}

\begin{document}

\title{AnomalyR1: A GRPO-based End-to-end MLLM for Industrial Anomaly Detection}

\author{Yuhao Chao \quad Jie Liu\thanks{Corresponding author} \quad Jie Tang \quad Gangshan Wu \\
State Key Laboratory for Novel Software Technology, Nanjing University, Nanjing 210023, China\\
{\tt\small 221240013@smail.nju.edu.cn \quad \tt\small \{liujie, tangjie, gswu\}@nju.edu.cn} 
}

\maketitle

\begin{abstract}
Industrial Anomaly Detection (IAD) poses a formidable challenge due to the scarcity of defective samples, making it imperative to deploy models capable of robust generalization to detect unseen anomalies effectively. Traditional approaches, often constrained by hand-crafted features or domain-specific expert models, struggle to address this limitation, underscoring the need for a paradigm shift. We introduce AnomalyR1, a pioneering framework that leverages VLM-R1, a Multimodal Large Language Model (MLLM) renowned for its exceptional generalization and interpretability, to revolutionize IAD. By integrating MLLM with Group Relative Policy Optimization (GRPO), enhanced by our novel Reasoned Outcome Alignment Metric (ROAM), AnomalyR1 achieves a fully end-to-end solution that autonomously processes inputs of image and domain knowledge, reasons through analysis, and generates precise anomaly localizations and masks. Based on the latest multimodal IAD benchmark, our compact 3-billion-parameter model outperforms existing methods, establishing state-of-the-art results. As MLLM capabilities continue to advance, this study is the first to deliver an end-to-end VLM-based IAD solution that demonstrates the transformative potential of ROAM-enhanced GRPO, positioning our framework as a forward-looking cornerstone for next-generation intelligent anomaly detection systems in industrial applications with limited defective data.
\end{abstract}

\section{Introduction}
\label{sec:intro}

\begin{figure}
    \centering
    \includegraphics[width=\linewidth]{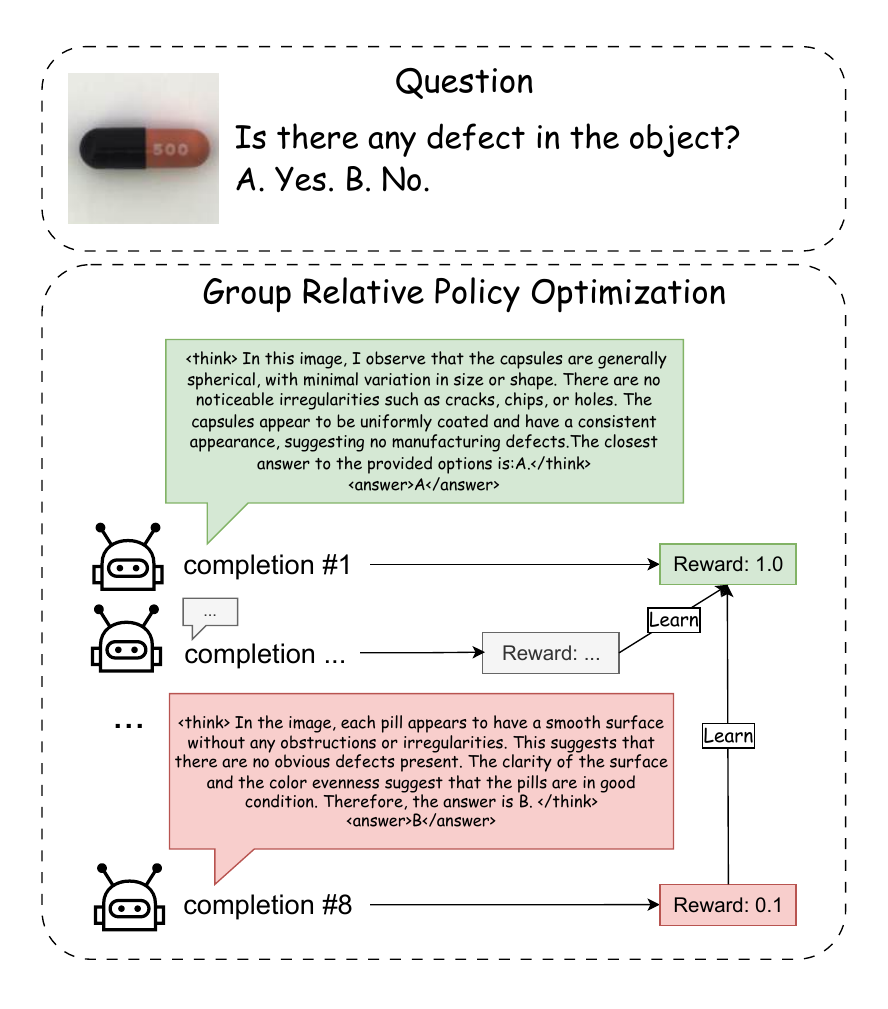}
    \caption{The process of group relative policy optimization: The model generates a set of responses for each prompt, scores them using a reward model, and updates its parameters based on the relative advantages within the group.}
    \label{fig1}
\end{figure}

In recent years, Multimodal Large Language Models (MLLMs) have emerged as a transformative force in artificial intelligence \cite{mllmSurvey}, with their ability to process and integrate diverse data modalities such as text, images, and sensor inputs. Models like CLIP \cite{clip}, GPT-4o \cite{hurst2024gpt4o}, and LLaMa \cite{touvron2023llama} leverage extensive pretraining in multimodal data sets to achieve impressive generalization and cross-modal reasoning capabilities. These strengths are particularly valuable for complex Industrial Anomaly Detection (IAD) tasks. IAD is a critical task in manufacturing that aims to identify defects or irregularities in complex systems, such as equipment failures or production line anomalies. Many excellent models \citep{guo2024dinomaly,zhu2024towards,liu2024dual, wang2023hybridfusion} can perform tasks well after specific training. Still, their lack of interpretability and insufficient multimodal capabilities reveal a deficiency in the potential for future, more extensive, and complex application scenarios.

Large-scale language models, such as MLLMs, are typically trained through resource-intensive processes, granting them immense potential for tasks like IAD \citep{xu2024customizing, zhang2024gpt4vad, gu2024filo, xu2025anomalyov, tian2024foct}. However, when applied to specific scenarios where training data is scarce, such as detecting rare defects in manufacturing, these models often require adaptation using limited resources. Their primary abilities are almost entirely determined by pre-training. While this integration emphasizes support for large language models, it significantly increases the system's complexity, reducing generalizability. We solve this challenge by driving the need to explore the ability of MLLMs to adapt to new tasks with minimal data. 

As analyzed in \autoref{tab:struct-comparison}, both expert models that are carefully designed for specific application scenarios and hybrid frameworks that integrate expert models with LLMs have their inherent limitations. They can only leverage part of the capabilities of LLMs and do not entirely focus on the multimodal abilities of MLLMs. This is somewhat inadequate in an era where large models are gradually unleashing increasingly powerful capabilities.

To address this, we introduce AnomalyR1, a novel model built on the foundation of VLM-R1 \cite{shen2025vlmr1}, significantly improved through our adaptation of Group Relative Policy Optimization (GRPO). GRPO optimizes strategy through intragroup comparison, normalizes reward calculation without a value function, reduces computational overhead, and is suitable for inference tasks. \autoref{fig1} briefly demonstrates its training process. This enhancement enables robust few-shot learning and requires only 2-5 images per class of industrial items to achieve adequate performance. GRPO, developed initially to improve mathematical reasoning in large-language models, optimizes performance within policy-driven frameworks by refining intrinsic reasoning capabilities. 

However, the original GRPO, tailored for mathematical tasks, struggled to generalize to IAD, particularly in critical cases where a disconnect emerged between the CoT reasoning and the final output. To overcome this, we developed the Reasoned Outcome Alignment Metric (ROAM) in AnomalyR1, specifically designed to tailor the reward mechanism during a few-shot learning on IAD datasets. This innovative approach iteratively guides the model through the reasoning process for industrial product images, incrementally improving its anomaly detection accuracy. Our contribution lies in this customized adaptation of GRPO, addressing its limitations in the IAD domain and enabling precise, context-aware reasoning. 

Evaluated on the latest multimodal IAD benchmark, MMAD \cite{Jiang2024MMAD}, AnomalyR1 delivers state-of-the-art (SOTA) performance, demonstrating its exceptional multimodal reasoning and generalization capabilities. This work underscores our advances in adapting GRPO, highlighting its potential beyond mathematical applications and solidifying AnomalyR1's position as a leading solution in industrial anomaly detection.

We summarize our \textbf{contributions} as follows:

(1) We apply MLLMs to IAD, especially in an end-to-end manner, which is a relatively new approach. This application distinguishes our work from traditional methods that rely on complex expert models or pre-trained vision-text encoders.  

(2) We propose AnomalyR1, built on VLM-R1 and enhanced by GRPO for few-shot learning. This training structure significantly improves IAD performance under data-scarce conditions. 

(3) We introduce the ROAM. This novel reward framework substantially elevates the performance of the original GRPO training method within the DeepSeekMath framework while also surpassing SFT techniques, especially in few-shot learning tasks for IAD.

(4) Our approach achieves SOTA performance on the latest multimodal IAD benchmark, demonstrating superior multimodal reasoning and generalization capabilities.

\begin{table}
    \centering
    \includegraphics[width=0.95\linewidth]{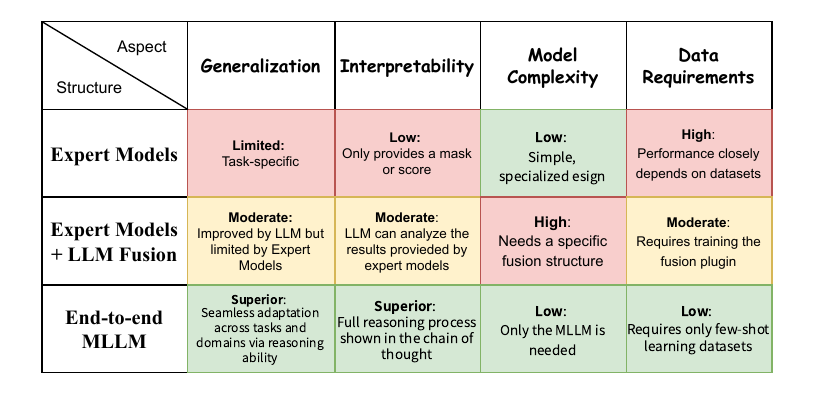}
    \caption{Comperision between different IAD model structures}
    \label{tab:struct-comparison}
\end{table}

\section{Related Work}

\subsection{Industrial Anomaly Detection}

Traditional industrial anomaly detection (IAD) methods primarily relied on classical image processing and statistical techniques to identify manufacturing defects \citep{liu2024deep, xie2024imiadindustrialimageanomaly}. With the emergence of few-shot learning (FSL), researchers began to address data scarcity by enabling models to generalize from a small number of labeled samples \citep{Wu_2021_ICCV, KAMOONA2024107706}. Meta-learning-based approaches, such as \citep{Wu_2021_ICCV, huang2022regad}, require large-scale meta-training to achieve adaptation, while others like PatchCore \cite{roth2022patchCore}, SPADE \cite{cohen2020SPADE}, and PaDim \cite{defard2021padim} utilize a minimal support set (e.g., 16 images) without specific FSAD optimization. Despite progress, these methods remain limited in efficiency and adaptability, especially in complex industrial settings.

To overcome the dependency on labeled samples, zero-shot anomaly detection (ZSAD) extends this line of research by leveraging large-scale pre-trained models. Notably, MAEDAY \cite{schwartz2024maeday} utilizes a masked autoencoder (MAE) \cite{he2022mae} for reconstruction-based anomaly localization, while CLIP-based methods such as WinCLIP \cite{jeong2023winclip}, AdaCLIP \cite{cao2024adaclip}, and AnomalyCLIP \cite{zhou2023anomalyclip} combine vision-language pretraining with feature matching. These approaches, summarized and compared in \autoref{tab:struct-comparison}, highlight the trade-offs between traditional expert systems, fusion-based designs, and modern end-to-end multimodal large models (MLLMs), particularly in terms of generalization, interpretability, and data requirements. However, unlike human experts, most models lack a contextual understanding of anomalies, often depending on mismatches in pre-trained patterns rather than reasoning about causal factors.

\subsection{MLLM for IAD}

The integration of MLLMs into IAD has become a vibrant area of research in recent years \citep{yang2025mllm_iad_survey, mllmSurvey}. Among the first contributions, Myriad \cite{li2023myriad} laid a foundational framework by combining large language models (LLM) with vision expert models. This pioneering approach established a classical structure that has significantly influenced subsequent studies exploring the integration of visual and linguistic processing. While Myriad's methodology was groundbreaking, its reliance on carefully curated vision expert models introduces complexity that can hinder scalability when applied to various industrial tasks.

Building on this groundwork, AnomalyGPT \cite{gu2023anomalyagpt} adapted MLLMs to interpret feature maps from expert models, achieving impressive zero-shot anomaly detection in industrial applications. While excelling without task-specific training, its reliance on external models limits its adaptability in dynamic settings. LogicAD \cite{jin2025logicad} and LogiCode \cite{zhang2024logicode} approached anomaly detection differently by framing it as a logical reasoning problem. They offered a unique perspective and proved most effective in scenarios where anomalies are clearly defined in rational terms, particularly in logic IAD datasets \cite{MvTecLOCO-dataset}. Echo \cite{chen2025can_mllm_iad} introduced a collaborative framework where specialized MLLMs work together, enhancing detection through system-level synergy, although it did not fully train MLLMs for IAD. \cite{zhang2025eiad} use SFT that cannot perform excellent growth in the MMAD benchmarks.

\subsection{Reinforcement Learning on Few-shot Data}

With the widespread adoption and development of large-scale models, significant efforts have been made to enhance the end-application capabilities of base models through fine-tuning. A classic technique, Supervised Fine-Tuning (SFT), has been widely applied, particularly in frameworks such as Explainable Industrial Anomaly Detection (EIAD) \cite{zhang2025eiad}. However, in industrial anomaly detection, SFT exhibits notable limitations, including its reliance on extensive labeled datasets, which are often scarce due to the rarity and diversity of anomalies, and its lack of interpretability, typically offering only pixel-level anomaly scores without deeper insights into anomaly categories or causes. Although RL methods, such as Proximal Policy Optimization (PPO) \cite{schulman2017ppo}, Direct Preference Optimization (DPO) \cite{rafailov2023dpo}. The recently proposed Group Relative Policy Optimization (GRPO) by the DeepSeekMath team has shown remarkable success in other complex reasoning tasks, such as mathematical problem-solving. Still, their application in industrial anomaly detection remains unexplored, particularly within MLLMs. Our research aims to address this gap by investigating the potential of these advanced RL techniques to enhance multimodal reasoning and interpretability in industrial anomaly detection.

In practical industrial settings, challenges such as limited computational resources and sparse sample data, driven by the unique characteristics of application scenarios, further complicate model training. As a result, improving model capabilities through RL with minimal samples has become a prominent research focus. Studies such as \citep{zhou2023lima, kirstain2021few, chen2023maybe05} have demonstrated the potential of few-shot learning approaches. By building on these insights, we introduce advanced RL techniques, including GRPO, in our research on Industrial Anomaly Detection. This work provides an end-to-end solution tailored to the domain and validates the effectiveness of RL-based fine-tuning in data-scarce environments. By addressing key challenges such as limited data and the need for interpretability, our framework enhances multimodal reasoning and boosts model performance in industrial anomaly detection.

\section{Method}

\subsection{Overview of AnomalyR1}

\begin{figure*}
    \centering
    \includegraphics[width=0.9\linewidth]{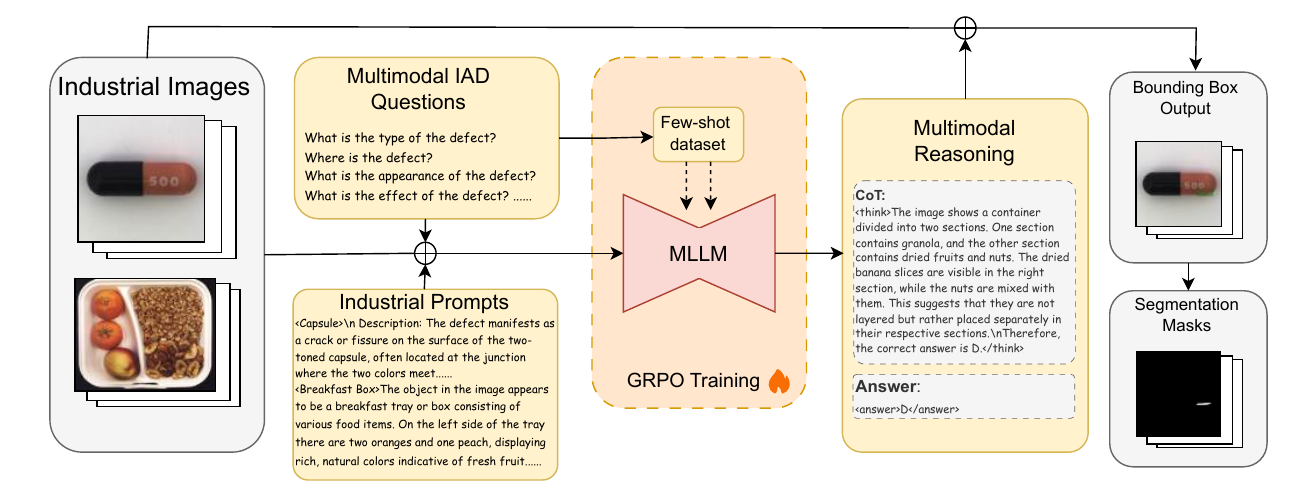}
    \caption{The structure of AnomalyR1, the model enhanced by GRPO training, is able to finish the whole end-to-end process, which shows the new paradigm for IAD tasks.}
    \label{fig:e2e-process}
\end{figure*}

Our approach fundamentally diverges from prior research by introducing a fully end-to-end architecture that maximizes the potential of Multimodal Large Language Models (MLLMs) for Industrial Anomaly Detection (IAD). \autoref{fig:e2e-process} shows the whole process of an end-to-end IAD MLLM system. Unlike traditional methods, our framework leverages MLLM to handle the entire anomaly detection pipeline without external components. The trade-off, however, is the requirement for few-shot learning strategies, which in turn endow our framework with strong generalization and portability across diverse IAD tasks. This holistic integration ensures that the model autonomously interprets multimodal inputs and delivers actionable outputs, eliminating the need for fragmented processing steps. We require the MLLM to provide detailed positional information to achieve precise anomaly localization by instructing the model to output anomaly locations directly in JSON format, yielding bounding box coordinates as a practical and resource-efficient solution.

To further refine this output for compatibility with datasets requiring RGB masks common in IAD benchmarks, we employ the state-of-the-art Segment Anything Model 2 (SAM2) \cite{sam2} to convert bounding boxes into pixel-level masks. This conversion step, while external, is a post-processing enhancement rather than a core component of our end-to-end pipeline, ensuring the MLLM remains the primary driver of anomaly detection. At this level, bounding box outputs capture the model's ability to localize anomalies accurately, aligning with practical industrial needs where high-level positional awareness often outweighs fine-grained detail. This strategic choice underscores our focus on developing a streamlined, adaptable framework that provides reliable results under constrained conditions.

Training MLLM to transition from textual descriptions to JSON-formatted bounding-box annotations required a robust dataset and methodology. We utilized the Ref-COCO \cite{kazemzadeh2014referitgame} dataset, renowned for its detailed object localization annotations, to guide the model in aligning visual inputs with precise positional outputs. The training process adopted the classic GRPO \cite{DeepSeekMath} strategy, too. In our adaptation, GRPO facilitates the model’s ability to iteratively refine its understanding of anomaly locations by optimizing policy-driven decisions across few-shot examples. This approach ensures that the model learns to generate structured outputs effectively and reinforces its capacity to generalize from limited data, a critical advantage in IAD, where defective samples are scarce. By integrating these elements, our end-to-end implementation demonstrates a cohesive and innovative application of MLLMs, setting a new standard for efficiency and performance in industrial anomaly detection.

This section introduces the Group Relative Policy Optimization (GRPO) methods in \autoref{sec:group-relative-policy-optimization}. Next, we present our Reasoned Outcome Alignment Metric (ROAM) optimization method for GRPO in the context of In-Context Adaptive Few-shot Learning (IAD) in \autoref{sec:reasoned-outcome-alignment-metric}. We then discuss the basic settings of our models in \autoref{sec:base-model-and-prompt-template}. Finally, we detail the preparation of few-shot data in \autoref{sec:few-shot-data-preparatioin}.

\subsection{Group Relative Policy Optimization}
\label{sec:group-relative-policy-optimization}

In reinforcement learning, the interaction between an agent and its environment is modeled through a sequence of states, actions, and policies. At each \textbf{time step $t$}, the agent observes the \textbf{environment's state $s_t$}, which encapsulates preprocessed information such as the agent's position or velocity, or even error metrics like speed deviation. Based on this state, the agent selects an action $a_t$, which represents the operation executed in the environment, such as moving forward or jumping. The decision-making process is governed by the \textbf{policy $\pi(a_t \mid s_t)$}, which defines the conditional probability of taking action $a_t$ given the state $s_t$. This policy can either be stochastic, outputting a probability distribution over possible actions, or deterministic, directly mapping the state to a specific action.
A key concept in policy optimization, particularly in Proximal Policy Optimization (PPO), is the policy ratio, which measures the relative change between the new policy and the old policy. This ratio is critical for ensuring stable updates during training. The \textbf{policy ratio $r_t(\theta)$} is defined as:
\begin{equation}
r_t(\theta) = \frac{\pi_\theta(a_t \mid s_t)}{\pi_{\theta_{\text{old}}}(a_t \mid s_t)},
\end{equation} where $\pi_\theta(a_t \mid s_t)$ represents \textbf{the new policy parameterized by $\theta$}, and $\pi_{\theta_{\text{old}}}(a_t \mid s_t)$ denotes the \textbf{old policy before the update}. This ratio balances the exploration of new strategies while preventing large, destabilizing policy updates, thereby improving the robustness of the learning process.

Originally developed to improve mathematical reasoning in large language models (LLMs), as shown in DeepSeekMath \cite{DeepSeekMath}, Group Relative Policy Optimization (GRPO) has been adapted in our research for few-shot Industrial Anomaly Detection (IAD) and multimodal understanding. We refined its reinforcement learning framework to thrive in data-scarce settings and integrate diverse modalities like images and textual logs, advancing its use in industrial applications.

Unlike conventional reinforcement learning algorithms such as PPO \cite{schulman2017proximal}, which typically rely on well-defined scalar reward signals and large-scale datasets, GRPO offers two crucial advantages particularly suited to IAD scenarios: (1) its group-wise relative advantage estimation allows for meaningful learning signals even in the absence of task-specific scalar rewards, and (2) its regularization via reference policies helps to stabilize updates under extremely limited supervision. These features make GRPO inherently robust to the challenges of IAD, where anomaly labels are rare, ambiguous, and often multimodal in nature. Therefore, GRPO not only addresses the scarcity of annotations but also promotes generalization across diverse industrial contexts.

To this end, we introduce a tailored GRPO framework for multimodal anomaly detection, with a reward mechanism that balances task performance and policy stability. In PPO, the reward function is defined as follows:
\begin{equation}
r_{\tau} = r_\phi(q, o_{\leq t}) - \beta \log \frac{\pi_{\theta}(o_t \| q, o_{<t})}{\pi_{ref}(o_t \| q, o_{<t})},    
\end{equation}
where $r_\phi(q, o_{\leq t})$ is the task-specific reward based on the observation sequence. At the same time, the second term regularizes the policy by keeping it close to a reference policy, mitigating overfitting in few-shot scenarios. $\pi_{\theta}$ is the last policy. But we need the reference policy ($\pi_{ref}$), which is always trained by Supervised Fine-tuning (SFT), which is not useful in IAD benchmarks (we will show it in \autoref{sec:multimodal-industrial-anomaly-detection}). It is also hard to prepare the task-specific reward on the IAD dataset, owing to our limited dataset.

GRPO sample a group of outputs $\{o_1, o_2, \dots, o_G\}$ from the old policy $\pi_{\theta_{old}}$. Our target is to maximize the following objective.

\begin{align}
\small
\mathcal{J}_{\text{GRPO}}(\theta) &= \mathbb{E}\left[ q \sim P(Q), \{O_i\}_{i=1}^G \sim \pi_{\theta_{\text{old}}}(O|Q) \right] \notag \\
&\quad \frac{1}{G} \sum_{i=1}^G \frac{1}{|O_i|} \sum_{t=1}^{|O_i|} \left\{ M_{i,t} - \beta \mathbb{D}_{\text{KL}}[\pi_{\theta}||\pi_{\text{ref}}] \right\},
\label{eq:grpo}
\end{align}
where:
\begin{align}
M_{i,t} &= \min [ \frac{\pi_{\theta}(O_{i,t}|Q, O_{i,<t})}{\pi_{\theta_{\text{old}}}(O_{i,t}|Q, O_{i,<t})} \hat{A}_{i,t},  \notag \\ 
& \quad \text{clip} \left( \frac{\pi_{\theta}(O_{i,t}|Q, O_{i,<t})}{\pi_{\theta_{\text{old}}}(O_{i,t}|Q, O_{i,<t})}, 1-\epsilon, 1+\epsilon \right) \hat{A}_{i,t} ].
\end{align}

In the equations above, the $\epsilon$ and $\beta$ are hyperparameters. And $\hat{A}_{i,t}$ is the advantage calculated based on the relative rewards of the outputs inside each group. The clip function constrains the upper and lower bounds of $\hat{A}_{i,t}$, preventing excessive perturbations. GRPO uses KL divergence and omits the more complex calculation for $\hat{A}_{i,t}$.

Introducing the KL penalty into the loss function can enhance the model's capability to understand and reason for IAD while preventing the model from deviating too far from the original distribution, thereby improving training stability. And we estimate the KL penalty as follows:
\begingroup
\setlength{\belowdisplayskip}{3pt}
\begin{align}
\mathbb{D}_{KL} \left[ \pi_{\theta} || \pi_{ref} \right] = \frac{\pi_{ref} (o_{i,t} | q, o_{i,<t})}{\pi_{\theta} (o_{i,t} | q, o_{i,<t})} \\ - \log \frac{\pi_{ref} (o_{i,t} | q, o_{i,<t})}{\pi_{\theta} (o_{i,t} | q, o_{i,<t})} - 1.
\end{align}
\endgroup

\subsection{Reasoned Outcome Alignment Metric}
\label{sec:reasoned-outcome-alignment-metric}

We propose an advanced reward framework, termed the \textbf{Reasoned Outcome Alignment Metric (ROAM)}, designed to comprehensively evaluate model performance by integrating both the reasoning process's fidelity and the final answer's precision. Drawing inspiration from methodologies such as DeepSeekMath, ROAM encapsulates two critical dimensions: a \textit{process coherence score}, which quantifies the alignment of the model's reasoning with the expected solution trajectory, and an \textit{outcome congruence score}, which assesses the accuracy of the final answer against the ground truth. This dual-component structure ensures that the model is incentivized to produce correct and interpretable outputs, a pivotal requirement for high-stakes industrial applications where transparency is paramount.

The ROAM framework operates by taking as input the model's reasoning trace, denoted \( R_\mu \), and its final output, denoted \( O_\nu \), and comparing these against the reference solution, comprising the ground truth reasoning \( R_\gamma \) and the ground truth answer \( A_\gamma \). This detailed evaluation is particularly crucial for IAD tasks. IAD presents unique challenges: datasets are often highly unbalanced, anomalies can be subtle and varied, and the tasks themselves are inherently tricky, making it hard for models to learn effectively using classical GRPO methods. Unlike mathematical problems with clearly defined solution paths and difficulty metrics, IAD necessitates guiding the model to \textit{reason} about visual deviations and articulate \textit{why} a sample is anomalous instead of guessing a random choice. ROAM employs a lightweight semantic segmentation module to bridge this gap and enable practical training guidance. This module parses the model's potentially unstructured chain-of-thought content and answer format using sophisticated heuristic matching strategies, allowing for precise monitoring and control. This mechanism is vital for mitigating deviations from desired reasoning paths (e.g., correctly localizing and characterizing a defect) and enhancing training efficacy specifically for IAD's complex, reasoning-intensive nature.

The reward function is constructed as a generalized framework, eschewing fixed numerical assignments in favor of a modular and abstract formulation. Specifically, ROAM is expressed as:
\begin{equation}
\mathcal{R}(R_\mu, O_\nu, R_\gamma, A_\gamma) = \Phi(R_\mu, R_\gamma) + \Psi(O_\nu, A_\gamma, R_\mu),
\end{equation} 
where $ \Phi(R_\mu, R_\gamma) $ represents the \textbf{process alignment function}, evaluating the similarity or equivalence between the model’s reasoning $R_\mu$ and the ground truth reasoning $R_\gamma$. This function captures the quality and coherence of the reasoning process, potentially ranging from null or absent reasoning to full alignment with the expected solution steps. $\Psi(O_\nu, A_\gamma, R_\mu)$: Denotes the \textbf{outcome-context synergy function}, evaluating the correctness of the final answer $O_\nu$ relative to the ground truth answer $A_\gamma$, while also considering the contextual influence of reasoning $R_\mu$. This term introduces flexibility to reward partial correctness or consistency between reasoning and output, even when deviations from the ground truth occur.

The formulation of ROAM integrates the process alignment function $\Phi(R_\mu, R_\gamma) $ and the outcome-context synergy function  $\Psi(O_\nu,$\\$ A_\gamma, R_\mu)$ to enable a sophisticated evaluation. $ \Phi $ quantifies the fidelity of the model’s reasoning $ R_\mu $ against the reference $ R_\gamma $, promoting step-by-step alignment, while $ \Psi $ assesses the final answer’s correctness $ O_\nu $ relative to the ground truth $ A_\gamma $, considering the reasoning context $ R_\mu $. Both components are designed for adaptability; for instance, $ \Phi $ can be implemented using methods ranging from simple indicator functions for exact matches to nuanced similarity metrics offering partial credit, and its contribution diminishes when reasoning is absent ($ R_\mu = \emptyset $). Similarly, $ \Psi $ can incorporate conditional logic to modulate rewards, accommodating scenarios such as correct answers without reasoning or recognizing internal consistency between reasoning and outcome, even if incorrect. 

The ROAM offers a robust and flexible framework for model evaluation, advancing beyond simplistic accuracy-based rewards. By integrating process coherence and outcome congruence through the abstract functions $\Phi$ and $\Psi$, ROAM provides a sophisticated tool for training models that excel in both performance and interpretability, as demonstrated in the workflow of our semantic segmentation strategy. This design positions ROAM as the cornerstone for next-generation model optimization in interpretability-critical domains.

\subsection{Base Model and Prompt Template}
\label{sec:base-model-and-prompt-template}

In this work, we select Qwen2.5VL-3b \cite{bai2025qwen25vl} as our base model, a state-of-the-art multimodal large language model (MLLM) capable of processing textual and visual inputs. This choice is motivated by its instruct variant, which is fine-tuned for instruction-following tasks, making it ideal for handling complex, multi-step queries. Our application in industrial anomaly detection requires the model to identify anomalies in images and provide a transparent reasoning process, as understanding the decision-making steps is as critical as the detection itself. To meet this requirement, we incorporate Chain of Thought (CoT) reasoning, which enables the model to explicitly break down its decision-making process, thereby enhancing interpretability and trust in the system.

To effectively guide the model in performing anomaly detection while articulating its reasoning, we design a structured prompt template that ensures accuracy and explainability. The template is defined as follows:

\begin{verbatim}
"A conversation between User and Assist
ant. The user asks a choice question an
d the Assistant solves it. The assista-
nt first think about the reasoning pro-
cess in the mind and then provides the 
user with the answer."
"Respond with your reasoning in <think>
</think> tags "
"followed by a single letter answer in 
<answer> </answer> tags."
\end{verbatim}

This prompt structure is specifically crafted to align with the needs of IAD. It instructs the model to first reason through the problem, such as analyzing an image of a machine part for defects like cracks or irregularities within <think> tags, and then deliver a concise decision in <answer> tags. By combining the capabilities of Qwen2.5VL-Instruct-3b with this prompt design, we establish a robust framework that ensures precision in anomaly detection and clarity in explaining the reasoning process, both essential for industrial applications. Due to resource constraints, our experiments were conducted entirely using the 3b small model for validation. However, according to widely accepted scaling laws, larger models are likely to exhibit even stronger performance under our anomalyR1 method. Similarly, larger commercial models are expected to demonstrate significantly enhanced capabilities in IAD application scenarios.

\subsection{Few-shot Data Preparation}
\label{sec:few-shot-data-preparatioin}

In our few-shot learning approach, we trained the model on 600 images sourced from traditional industrial datasets \citep{MvTecLOCO-dataset, Bergmann_2019_CVPR_MVTecAD, zou2022spot_VisA, zhang2024pku_GoodsAD}, inspired by similar sample settings in the MMAD \cite{Jiang2024MMAD} dataset. By adapting problems from MMAD datasets, we enabled the model to quickly familiarize itself with the problem format and focus on improving its CoT reasoning and the relevance of its responses. This strategy not only demonstrates the model's ability to generalize effectively from a minimal yet representative training set crucial in industrial anomaly detection where defective samples are scarce but also highlights the practical utility of our method in real-world applications.

\section{Experiments}

\begin{table*}[t]
\vspace{-0.8em}
\centering
\setlength\tabcolsep{3pt}
\caption{Performance comparison of both proprietary and open-source MLLMs in MMAD with the standard 1-shot in-context learning setting. Anomaly Discrimination uses the average accuracy of normal and abnormal categories. (*For the methods not supporting multi-image input, they only receive the image of the question itself)} 
\label{mmad-performance-comparison}
\resizebox{\linewidth}{!}{
\begin{tabular}{c|c|c|cccc|cc|c}
\toprule\toprule
                         &                         & Anomaly        & \multicolumn{4}{c|}{Defect}                             & \multicolumn{2}{c|}{Object} &                           \\\cmidrule(r){3-9}
\multirow{-2}{*}{Model} & \multirow{-2}{*}{Scale} & Discrimination & Classification & Localization & Description & Analysis & Classification  & Analysis & \multirow{-2}{*}{Average} \\\midrule
Random Chance         & -                       & 50.00          & 25.00          & 25.00        & 25.00       & 25.00    & 25.00           & 25.00    & 28.57                     \\ \midrule
\rowcolor{myblue} 
Human (expert)          & -                       & 95.24           & 75.00           & 92.31         & 83.33     & 94.20        & 86.11            & 80.37     & 86.65                     \\
\rowcolor{myblue} 
Human (ordinary)        & -                       & 86.90          & 66.25          & 85.58        & 71.25    & 81.52       & 89.58           & 69.72    & 78.69  
\\ \midrule
\rowcolor{mygray} 
Claude-3.5-sonnet        & -                       & 60.14           & 60.14           & 48.81         & 67.13        & 79.11     & 85.19            & 79.83     & 68.36                      \\
\rowcolor{mygray} 
Gemini-1.5-flash         & -                       & 58.58          & 54.70          & 49.10        & 66.53       & 82.24    & 91.47           & 79.71    & 68.90                     \\
\rowcolor{mygray} 
Gemini-1.5-pro           & -                       & \textbf{68.63}          & 60.12          & \textbf{58.56}        & 70.38       & 82.46    & 89.20           & 82.25    & 73.09                     \\
\rowcolor{mygray} 
GPT-4o-mini              & -                       & 64.33          & 48.58          & 38.75        & 63.68       & 80.40    & 88.56           & 79.74    & 66.29                     \\
\rowcolor{mygray} 
GPT-4o                   & -                       & \textbf{68.63}          & \textbf{65.80}          & 55.62        & \textbf{73.21}       & \textbf{83.41}    & \textbf{94.98}           & \textbf{82.80}    & \textbf{74.92}                     \\\midrule
AnomalyGPT \cite{gu2023anomalyagpt}          & 7B                      & 65.57          & 27.49          & 27.97        & 36.86       & 32.11    & 29.84           & 35.82    & 36.52                     \\
Qwen-VL-Chat \cite{bai2023qwen}       & 7B                      & 53.65          & 31.33          & 28.62        & 41.66       & 63.99    & 74.46           & 67.94    & 51.66                     \\
LLaVA-1.5  \cite{liu2024improved}         & 7B                      & 51.33          & 37.04          & 36.62        & 50.60       & 69.79    & 68.29           & 69.53    & 54.74                     \\
Cambrian-1*   \cite{tong2024cambrian}           & 8B                      & 55.60          & 32.53          & 35.39        & 43.46       & 49.14    & 78.15           & 67.22    & 51.64                     \\
SPHINX*  \cite{lin2023sphinx}        & 7B                      & 53.13          & 33.93          & 52.27        & 50.96       & 71.23    & 85.07           & 73.10    & 59.96                     \\
LLaVA-NEXT-Interleave \cite{li2024llavanextinter} & 7B                      & 57.64          & 33.79          & 47.72        & 51.84       & 67.93    & 81.39           & 74.91    & 59.32                     \\
InternLM-XComposer2-VL \cite{dong2024internlm}   & 7B                      & 55.85          & 41.80          & 48.27        & 57.52       & 76.60    & 74.34           & 77.75    & 61.73                     \\
LLaVA-OneVision  \cite{li2024llavaov}     & 7B                      & 51.77          & 46.13          & 41.85        & 62.19       & 69.73    & 90.31           & 80.93    & 63.27                     \\
MiniCPM-V2.6   \cite{yao2024minicpm}    & 8B                      & 57.31          & 49.22          & 43.28        & 65.86       & 75.24    & \textbf{92.02}           & 80.80    & 66.25                     \\
InternVL2  \cite{chen2024internvl}         & 8B                      & 59.97          & 43.85          & 47.91        & 57.60       & 78.10    & 74.18           & 80.37    & 63.14                     \\
LLaVA-1.5  \cite{liu2023llava} & 13B                     & 49.96          & 38.78          & 46.17        & 58.17       & 73.09    & 73.62           & 70.98    & 58.68                     \\
LLaVA-NeXT  \cite{liu2024llavanext}    & 34B                     & 57.92          & 48.79          & 52.87        & 71.34       & 80.28    & 81.12           & 77.80    & 67.16                     \\
InternVL2    & 76B                     & \textbf{68.25}          & 54.22          & 56.66       & 66.30       & 80.47    & 86.40           & 82.92    & 70.75                    \\
Qwen2.5-VL \cite{bai2025qwen25vl} (base) & \textbf{3B}                  & 51.10          & 41.07          & 44.87        & 61.43       & 75.99    & 86.54           & 79.58    & 62.94                     \\
Qwen2.5-VL (sft) & \textbf{3B}                  & 53.36          & 41.51          & 49.73        & 62.84       & 75.58    & 86.39           & 79.58    & 64.13                     \\
\textbf{AnomalyR1 (ours) }        & \textbf{3B}                      & 60.62          & \textbf{63.56}          & \textbf{70.14}        & \textbf{80.47}       & \textbf{85.28}    & \textbf{92.48}           & \textbf{86.15}    & \textbf{76.96}                     \\
\bottomrule\bottomrule
\end{tabular}
}
\end{table*}

\subsection{Trainning Settings}

We trained our base model using four Nvidia RTX A6000 48G GPUs. In the training process, we employed a batch size of 8, meaning that the model processes eight input samples simultaneously in each step, and we train our model for 1000 steps. Concurrently, the model output is structured to generate eight answers per group. This grouping indicates that each of the eight input samples in a batch corresponds to one answer, and these eight answers are processed as a unit. We use flash attention 2 \cite{dao2023flashattention2} and deepspeed \cite{Rasley2020deepspeed} to accelerate our training process and save our GPU memory usage. To maintain an end-to-end process, we did not apply any techniques to enhance the accuracy of our answers through image processing. In contrast, to train the model within limited resources, we compressed some ultra-high-resolution industrial images to $1000\times800$ pixels, which might have even worsened our results. By aligning the batch size with the answer grouping, this design maximizes computational resource utilization in each iteration while enhancing the model's ability to discern subtle differences between samples.

\subsection{Evaluation}

Since our framework introduces a new perspective on the task, we include comparisons with the most relevant existing models, although the availability of directly comparable baselines remains limited. In our study, we apply the MMAD \cite{Jiang2024MMAD} dataset as our evaluation benchmark. It has seven subtasks: 
\textbf{Anomaly Discrimination Detection:} – a binary classification task asking whether a sample has defects, testing the model's ability to detect anomalies; \textbf{Defect Classification:} – determining the type of defect, which assesses both anomaly detection and the model's knowledge of industrial defect categories; \textbf{Defect Localization:} – requiring the model to identify the defect's exact location, with standardized textual descriptions used in place of direct mask outputs;  \textbf{Defect Description:} – describing defect characteristics (e.g., size, color), simulating real-world defect analysis; \textbf{Defect Analysis:} – analyzing the potential impact of the defect on the product to assess its severity; \textbf{Object Classification:} – categorizing industrial products to help identify anomalies based on an understanding of normal product characteristics; \textbf{Object Analysis:} – questioning the composition, position, and function of the product to assess detailed understanding. 

These tasks are designed to mimic the full spectrum of duties that a quality inspector performs. The evaluation program randomizes the answer choices to avoid the probability of overfitting. To ensure the generalization of our model, we transfer another MVTec3D-AD dataset \cite{bergmann2021mvtec3dad} to the same format with detailed questions about the anomaly within the images. We keep the zero-shot settings in this experiment.

In MMAD, by designing the multiple choice questions related to multimodal understanding, four classic industrial anomaly detection datasets were improved. The evaluation of the final performance is based on accuracy, and it uses the macro-average method. This can effectively balance the different sample sizes of different datasets and better evaluate the overall capability of the model. Secondly, this can highlight the model's issues on a certain type of problem (such as the usually most challenging defect detection). It can very effectively increase researchers' attention to generalizability and satisfy the basic demands of research in the IAD field.

\subsection{Multimodal Industrial Anomaly Detection} 
\label{sec:multimodal-industrial-anomaly-detection}

The MMAD benchmark is designed to evaluate the performance of MLLMs on multimodal IAD tasks. However, most MLLMs do not know some basic rules behind IAD tasks. As part of this setup, the benchmark provides each model with a single reference image alongside the test image to support in-context learning (ICL), enabling the MLLMs to focus better on the detection objects. However, recognizing that some open-source MLLMs do not support multiple image inputs, the evaluation is adapted for these models by using only the test image, thus maintaining a zero-shot approach. In our experiments on this benchmark, our proposed model achieved state-of-the-art results, as detailed in \autoref{mmad-performance-comparison}, surpassing existing methods and delivering a performance improvement of over $14\%$ in the primary evaluation metric compared to our base model. Notably, our lightweight 3B model demonstrated performance comparable to or even exceeding that of larger models, showcasing the strength and efficiency of our approach. Furthermore, its capabilities in ‘Object Classification’ and ‘Object Analysis’ even surpass those of human experts, highlighting the remarkable potential of our model in complex multimodal understanding tasks.

\subsection{Comparative Study}

\begin{table}[h]
\centering
\caption{Comparison of Qwen2.5VL under 1-shot and 2-shot settings across different datasets}
\label{tab:icl-result}
\resizebox{\linewidth}{!}{
\begin{tabular}{lccccc}
\toprule
\textbf{Setting} & \textbf{DS-MVTec} & \textbf{MVTec-LOCO} & \textbf{VisA} & \textbf{GoodsAD} & \textbf{Average} \\
\midrule
1-shot  & 72.03 & 58.44 & 63.64 & 63.37 & 64.37 \\
2-shot  & 71.78 & 57.53 & 63.27 & 62.41 & 63.75 \\
AnomalyR1 & 81.99 & 70.55 & 76.69 & 78.60 & 76.96 \\
\bottomrule
\end{tabular}
}
\end{table}

In the application scenarios of IAD, there are different research approaches to a small number of sample images. In MMAD, the authors employ 1-shot and 2-shot settings; however, their implementation departs from the standard few-shot learning protocol. Rather than sampling fixed support examples, MMAD retrieves the most similar image from the training set. This retrieval-based strategy effectively harnesses a more extensive collection of data than is typically permitted in conventional few-shot learning, as it taps into the entire training set to select the most informative example. While this may offer an advantage, it raises concerns about fairness compared to methods like ICL that adhere to random sampling. In contrast, our proposed framework achieves comparable or superior outcomes, showcasing its efficiency under similar constraints.

To evaluate this retrieval-based setup, we conducted experiments with the base model under both 1-shot and 2-shot configurations. As reported in \autoref{tab:icl-result}, the 1-shot setting yields a slight improvement over the baseline, reflecting a modest gain in model capacity. However, the 2-shot setting underperforms compared to the 1-shot scenario, implying that additional retrieved images may disrupt the model's ability to integrate multimodal information effectively. These results reveal the inefficiencies of excessive data reliance in ICL-like approaches. Conversely, our proposed few-shot learning framework excels with the same limited data, delivering robust and efficient performance across configurations, thus outperforming traditional ICL methods in leveraging constrained support examples.

\begin{table*}[h]
\centering
\caption{Comparison of AnomalyR1 and Qwen2.5VL (base) across various categories in the anomaly understanding task of the MVTec3D-AD dataset (zero-shot settings)}
\resizebox{\linewidth}{!}{
\begin{tabular}{lccccccccccc}
\hline
 & foam & cookie & rope & peach & tire & potato & bagel & carrot & dowel & cable\_gland & \textbf{Average} \\
\hline
Qwen2.5VL \cite{bai2025qwen25vl} & $42.81$ & $59.58$ & $46.39$ & $56.01$ & $51.01$ & $55.30$ & $64.63$ & $55.78$ & $40.38$ & $52.01$ & \textbf{$52.11$} \\
AnomalyR1 & $47.81$ & $62.67$ & $46.33$ & $65.34$ & $55.24$ & $58.63$ & $65.77$ & $63.76$ & $43.03$ & $59.87$ & \textbf{$56.85$} \\
\textit{Promotion} & \textbf{$+5.00$} & \textbf{$+3.09$} & \textbf{$-0.06$} & \textbf{$+9.32$} & \textbf{$+4.24$} & \textbf{$+3.33$} & \textbf{$+1.14$} & \textbf{$+7.99$} & \textbf{$+2.64$} & \textbf{$+7.86$} & \textbf{$+4.74$} \\
\hline
\end{tabular}
}
\label{tab:bracket_comparison}
\end{table*}

\subsection{Ablation Study}

\begin{table}[ht]
\centering
\caption{Ablation experiment of ROAM on Defect Detection}
\resizebox{\linewidth}{!}{
\begin{tabular}{lccccc}
    \toprule
    & DS-MVTec & MVTec-LOCO & VisA & GoodsAD & Average \\
    \midrule
    Base Model     & 52.74 & 50.05 & 51.44 & 50.16 & 51.10 \\
    Classical GRPO & 60.11 & 49.96 & 49.95 & 50.55 & 52.64 \\
    GRPO with ROAM & 65.07 & 56.30 & 69.03 & 52.08 & 60.62 \\
    \bottomrule
\end{tabular}
}
\label{tab:grpo-ablation}
\end{table}

Our proposed ROAM markedly enhances the efficacy of GRPO, as evidenced by its superior performance over classical GRPO training in anomaly detection tasks. Evaluated after 1000 training steps across multiple datasets, ROAM achieves an average accuracy of 60.62 compared to 52.64 for classical GRPO, as detailed in \autoref{tab:grpo-ablation}. This improvement stems from ROAM's ability to mitigate the tendency of models to rely on guesswork during training, thereby fostering genuine reasoning and boosting detection accuracy. To operationalize ROAM, we adopted a hyperparameter configuration: a score of 0 is assigned when the reasoning process and answer are inconsistent; 0.05 is awarded when both a reasoning process and an answer are present, regardless of correctness; 0.1 is given when the reasoning process and answer are consistent; 0.8 is granted for a correct answer alone; and a full score of 1.0 is achieved when the reasoning process includes a derivation leading to the correct answer. This structured reward system ensures that both accuracy and coherent reasoning are prioritized.

Despite these gains, the overall anomaly detection scores remain relatively modest, underscoring the inherent difficulty of the task and the significance of our performance improvements. For example, in the GoodsAD dataset, the improvement from 50.55 to 52.08 may seem marginal. However, the baseline accuracy $50\%$ reflects in a large way incorrect guessing rather than reasoned inference, while our $52\%$ accuracy signifies a substantial enhancement in the reasoning capacity of the model. These results affirm ROAM's potential to advance few-shot learning in Industrial Anomaly Detection, particularly in data-scarce environments where interpretability and precision are critical.

\subsection{Generalization Study}

To validate the generalization of our few-shot learning GRPO results, which were derived from four classic industrial datasets, we adapted a well-established and challenging industrial anomaly detection (IAD) dataset, MVTec3D-AD \cite{bergmann2021mvtec3dad}. We utilized only its RGB inputs for a multimodal understanding and inference task, further increasing the task's complexity. Although the names appear similar, this dataset does not share any overlapping images with the dataset used in our few-shot learning process, which ensures that this experiment effectively evaluates the model's generalization ability. Following the setup of MMAD, we reformulated the understanding of image anomalies into multiple-choice questions for evaluation. In total, we designed four types of multiple-choice questions, encompassing anomaly detection, analysis of object categories, and analysis of anomaly types and locations.

As shown in \autoref{tab:bracket_comparison}, AnomalyR1 demonstrates comprehensive improvements across all categories compared to the baseline model in zero-shot settings, with particularly notable enhancements in the "peach," "carrot," and "cable gland" categories. These results,provide strong evidence supporting the effectiveness and generalization capability of our GRPO training approach.

\subsection{Qualitative Results}

\begin{figure}[h]
    \centering
    \includegraphics[width=\linewidth]{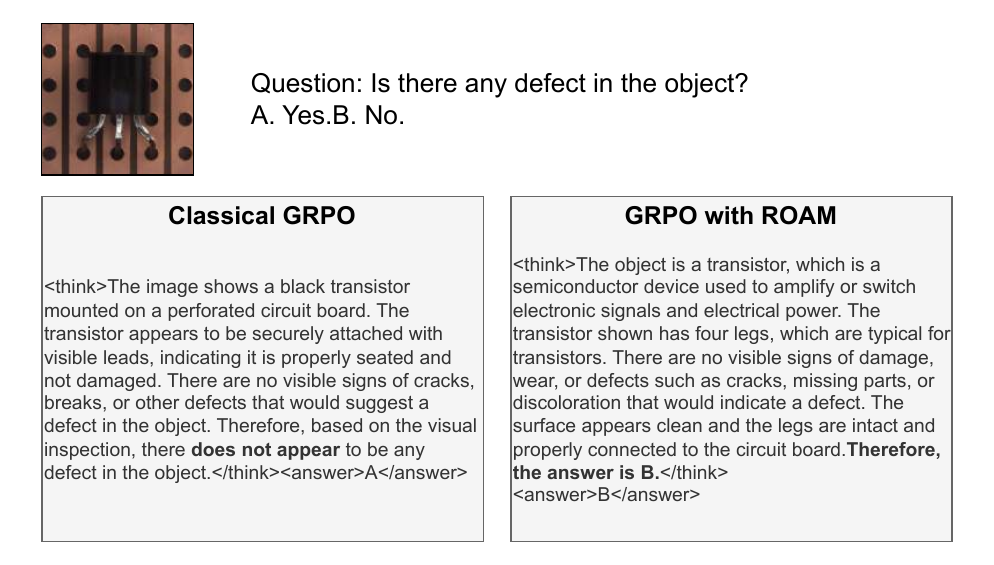}
    \caption{GRPO with ROAM (AnomalyR1) shows a better reasoning process and gives out the correct answer, while classical GRPO shows a conflict between CoT and answer.}
    \label{fig:qualitative-result}
\end{figure}

To further illustrate the effectiveness of our proposed ROAM optimization, we present a case study from an anomaly detection task in \autoref{fig:qualitative-result}. We compare the outputs of the model trained with the standard GRPO method and our ROAM-enhanced model. Although the GRPO-based model produces a seemingly correct reasoning trace, it ultimately provides an incorrect answer. This reflects the "guessing behavior" described in our ablation study, where the model generates plausible explanations but fails to verify its conclusions. In contrast, the ROAM-optimized model incorporates a self-verification process during reasoning, which leads to a more logically coherent answer. This improves the reliability of the inference process and enhances the model's capability to integrate multimodal information effectively.

\section{Conclusion}

Our research demonstrates that GRPO training significantly enhances performance in industrial anomaly detection by unlocking the reasoning potential of MLLMs, even when their scale appears limited. By leveraging few-shot learning and the ROAM methodology, the model can effectively self-improve using only externally provided correct answers. This innovative strategy not only boosts the model's capabilities but also highlights the latent strengths of smaller MLLMs in tackling complex reasoning tasks. Notably, our ROAM framework achieves remarkable performance improvements with only few-shot learning, suggesting strong potential for application in a broader range of domains. This demonstrates the versatility of our end-to-end approach and invites further validation in diverse real-world scenarios.

Future work could focus on enhancing MLLMs’ reasoning through more diverse and generalizable multimodal training datasets. Improving the GRPO reinforcement learning approach and exploring novel reinforcement methods may also boost few-shot learning performance.

{\small
\bibliographystyle{ieee_fullname}
\bibliography{anomalyr1}
}

\end{document}